\documentclass{article}

\PassOptionsToPackage{numbers, compress}{natbib}

\usepackage[preprint]{neurips_2023}

\usepackage[utf8]{inputenc} 
\usepackage[T1]{fontenc}    
\usepackage{hyperref}       
\usepackage{url}            
\usepackage{booktabs}       
\usepackage{amsfonts}       
\usepackage{nicefrac}       
\usepackage{microtype}      
\usepackage{xcolor}         

\usepackage{times}
\usepackage{latexsym}
\usepackage{enumitem}

\usepackage{graphicx}
\usepackage{amssymb}
\usepackage{comment}

\usepackage{caption}
\usepackage{subcaption}
\usepackage{booktabs}

\usepackage{multicol}
\usepackage{multirow}
\usepackage{array, tabularx,  ragged2e,  booktabs}
\usepackage{natbib}

\usepackage[utf8]{inputenc}
\usepackage{graphicx}
\usepackage{pgfplots}
\pgfplotsset{compat=1.14}

\usepackage{ctable}
\usepackage{wrapfig}
\usepackage{tikz}
\usetikzlibrary{positioning,arrows,trees,shapes,fit,shadows}

\usepackage{amsfonts}
\usepackage{ifthen}
\usepackage{booktabs}
\usepackage{amssymb}
\usepackage{supertabular}
\usepackage{array}
\usepackage{multirow}
\usepackage{fancyhdr}
\usepackage[normalem]{ulem}

\usepackage{amssymb}
\usepackage{supertabular}
\usepackage{array}
\usepackage{multirow}
\usepackage{fancyhdr}
\usepackage{psfrag}
\usepackage{amsmath}
\usepackage{hhline}
\usepackage{type1cm}
\usepackage{lettrine}
\usepackage[colorinlistoftodos,prependcaption,textsize=tiny]{todonotes}

\usepackage{algorithm}
\usepackage[noend]{algpseudocode}

\usepackage{xspace}

\usepackage{babel}
\usepackage[export]{adjustbox}
\usepackage{cleveref}

\usepackage{scrextend}

\crefformat{section}{\S#2#1#3} 
\crefformat{subsection}{\S#2#1#3}
\crefformat{subsubsection}{\S#2#1#3}

\usepackage{amsmath,amsfonts,bm}

\newcommand{\system}{{\texttt{System}}\xspace}
\newcommand{\user}{{\texttt{User}}\xspace}
\newcommand{\assistant}{{\texttt{Assistant}}\xspace}
\newcommand{\observation}{{\texttt{Observation}}\xspace}
\newcommand{\thought}{{\texttt{Thought}}\xspace}

\def\eqref#1{equation~\ref{#1}}

\def\1{\bm{1}}

\makeatletter
\newcommand{\xRightarrow}[2][]{\ext@arrow 0359\Rightarrowfill@{#1}{#2}}
\makeatother

\DeclareMathAlphabet{\mathsfit}{\encodingdefault}{\sfdefault}{m}{sl}
\SetMathAlphabet{\mathsfit}{bold}{\encodingdefault}{\sfdefault}{bx}{n}

\newcommand{\Ni}{({\em i})~}
\newcommand{\Nii}{({\em ii})~}
\newcommand{\Niii}{({\em iii})~}
\newcommand{\Niv}{({\em iv})~}

\usepackage{float}

\DeclareCaptionLabelFormat{andtable}{#1~#2  \&  \tablename~\thetable}

\usepackage{tikz-cd}
\usepackage{mathtools}
\usepackage{todonotes}
\usepackage{xcolor}

\definecolor{darkblue}{rgb}{0, 0, 0.5}
\hypersetup{
    colorlinks=true,
    linkcolor=darkblue,
    citecolor=darkblue,
    pdfpagemode=FullScreen,
    }

\title{\ourmodel: Towards Contextually Faithful LLMs}

\author{%
    Xuan-Phi Nguyen\thanks{Corresponding author: \href{mailto:xnguyen@salesforce.com}{xnguyen@salesforce.com}}~ \And Shrey Pandit~ \And Senthil Purushwalkam~ \And Austin Xu~~ \And Hailin Chen~ \And Yifei Ming~ \And Zixuan Ke~ \And Silvio Savarese~ \And Caiming Xong \And Shafiq Joty \\\\
    Salesforce AI Research 
}

\newcommand{\ourmodel}{{SFR-RAG}\xspace}
\newcommand{\oureval}{{ContextualBench}\xspace}

\begin{document}

\maketitle
\setcounter{footnote}{0} 

\begin{abstract}
Retrieval Augmented Generation (RAG), a paradigm that integrates external contextual information with large language models (LLMs) to enhance factual accuracy and relevance, has emerged as a pivotal area in generative AI. The LLMs used in RAG applications are required to faithfully and completely comprehend the provided context and users' questions, avoid hallucination, handle unanswerable, counterfactual or otherwise low-quality and irrelevant contexts, perform complex multi-hop reasoning and produce reliable citations. In this paper, we introduce \ourmodel, a small LLM that is instruction-tuned with an emphasis on  context-grounded generation and hallucination minimization. We also present ContextualBench, a new evaluation framework compiling multiple popular and diverse RAG benchmarks, such as HotpotQA and TriviaQA, with consistent RAG settings to ensure reproducibility and consistency in model assessments. Experimental results demonstrate that our \ourmodel-9B model outperforms leading baselines such as Command-R+ (104B) and GPT-4o, achieving state-of-the-art results in 3 out of 7 benchmarks in ContextualBench with significantly fewer parameters. The model is also shown to be resilient to alteration in the contextual information and behave appropriately when relevant context is removed. Additionally, the \ourmodel model maintains competitive performance in general instruction-following tasks and function-calling capabilities. 


\end{abstract}

\section{Introduction}\label{sec:intro}

\begin{figure}[h]
    \centering
    \includegraphics[width=0.5\linewidth]{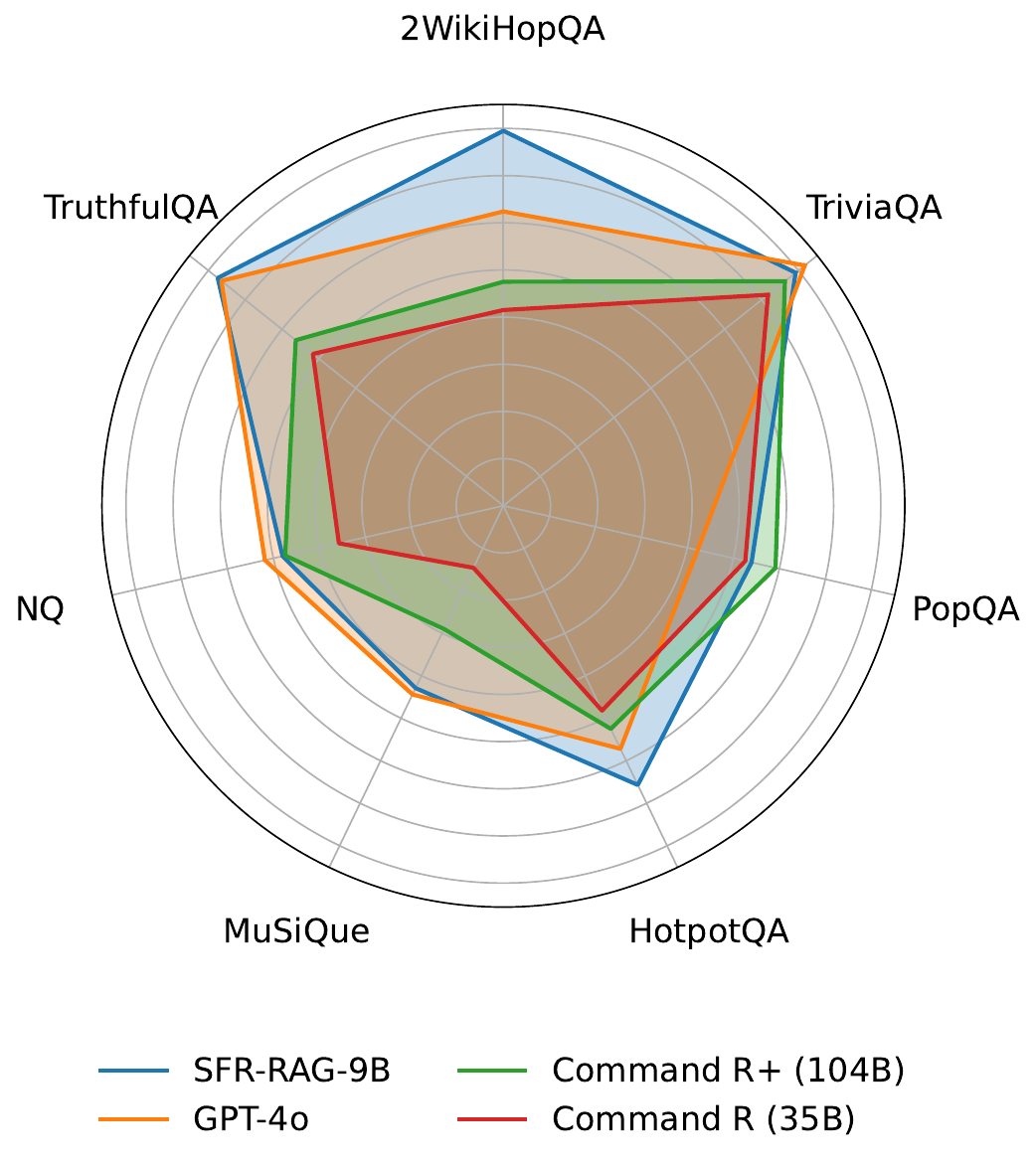}
    \small \caption{Our \ourmodel-9B model exhibits strong overall performance on an \oureval, our comprehensive evaluation suite of seven contextual tasks under a standardized setup. Notably, \ourmodel achieves state-of-the-art performance on three of seven tasks, with extremely competitive performance on the rest, despite having far fewer parameters than competitive baselines.}
    \label{fig:radar-plot}
\end{figure}

Retrieval Augmented Generation (RAG) has recently garnered significant attention as one of the most prominent areas of research in generative AI \citep{zhao2024retrieval,zhao-etal-2023-retrieving}, driven by the latest advancements in foundational large language models (LLMs) \citep{gpt3_brown2020language,lambda_google_thoppilan2022lamda,chatgpt,gpt4,llama2touvron2023llama,llama3_dubey2024llama,mistral7b_jiang2023mistral,qwen_bai2023}.
RAG frameworks are well-suited for solving knowledge-dependent problems or questions, where external contextual information is provided and the generated answer is expected to be \emph{factually grounded} on the contextual cues.
In practice, the RAG setup is designed such that a generator LLM works in tandem with a knowledge retriever. The retriever \citep{sfr_embedding_mistral,bge_embedding,bge_m3_embedding_chen2024bge,dense_retrieval_li2023making} is tasked with retrieving passages relevant to a given query from a database of documents (potentially the entire internet). The LLM interacts with users, formulates queries for the retriever to gather knowledge, and finally answers users' questions. 
To retrieve the most accurate context information, the retriever typically relies on an embedding model \citep{mteb_muennighoff2022mteb,sfr_embedding_mistral,bge_m3_embedding_chen2024bge,dense_retrieval_li2023making}, and optionally employs a re-ranker to get a refined list of context documents \citep{reranker_ma2023large}. Recent research has also led to the development of more sophisticated RAG frameworks \citep{selfrag_asai2023self,wikimultihopqa_constructing_ho2020,chain_of_knowledge_li2023chain,recomp_xu2023,bridging_retriever_llm_ke2024,weijia2023replug} that involve multiple inference steps to improve the reliability of answers.

In this work, we focus our efforts on the generator LLM component of the RAG framework. Traditional general-purpose LLMs trained for chat often struggle when directly applied to the RAG framework. This can be attributed to several potential factors, including:
\begin{itemize}[leftmargin=*]
    \item The knowledge in the context obtained from the retriever might conflict with the training data used for the LLM.
    \item The LLM is not trained to deal with conflicting or redundant facts from the retriever.
    \item In scenarios where the retrieved knowledge is insufficient, the LLMs revert to answering questions based on its training data. 
    \item It may also fail to provide adequate citations or to call appropriate functions and parameters to retrieve appropriate contexts in an agentic environment \citep{react_yao2022react}, in which the model may use provided functions or tools to perform tasks.
\end{itemize}
Recent attempts have focused on training LLMs specifically tuned to succeed in the RAG framework, such as \href{https://cohere.com/blog/command-r}{Command-R(+)} \citep{commandr} and \href{https://contextual.ai/introducing-rag2}{RAG-2.0} \citep{rag2}. 
Such RAG-specific LLMs not only serve as the foundation for generating up-to-date and factual AI responses, but also enable quick adoption in different domains, avoid the need to increase model capacity, context length or fine-tune an LLM on potentially proprietary data. 

In this work, we introduce \ourmodel\footnote{The model will be made available via API and later fully open-sourced.}, a 9-billion-parameter language model trained with a significant emphasis on reliable, precise and faithful contextual generation abilities specific to RAG and relevant agentic tasks. Beyond contextual tasks, \ourmodel\ is also trained to serve as a competitive AI assistant in regular tasks \citep{mmlu_hendryckstest2021,gsm8k_cobbe2021gsm8k}. We develop a comprehensive recipe, on both data synthesis and training procedures, to train the base LLM so that it is familiar and adaptable to diverse real-life RAG use cases. This includes precise factual knowledge extraction, distinguishing relevant against distracting contexts, citing appropriate sources along with answers, producing complex and multi-hop reasoning over multiple contexts, consistent format following, as well as refraining from hallucination over unanswerable queries. \ourmodel is also equipped with function calling and agentic abilities, which enable it to proactively search for knowledge from external tools, as well as conduct complex inference and reasoning strategies similar to Self-RAG \citep{selfrag_asai2023self}, ReAct \citep{react_yao2022react} and alike \citep{chain_of_knowledge_li2023chain,rankrag_yu2024rankrag,planrag_lee2024planrag,replugrag_shi2023replug}.

There is limited well established evaluation standards for measuring progress in contextual comprehension qualities of LLMs. It is worth noting that Command-R(+) \cite{commandr} and RAG-2.0 \cite{rag2} evaluated their proposed models on non-overlapping metrics \citep{hotpotqa_yang2018,truthfulqa_lin2021} with inconsistent or undisclosed setups, which causes difficulties in aligning results and comparisons across different studies. To reliably evaluate our \ourmodel model as well as other well-known baselines, in this work, we also introduce \oureval\footnote{\href{https://huggingface.co/datasets/Salesforce/ContextualBench}{https://huggingface.co/datasets/Salesforce/ContextualBench}}, which is a compilation of many popular RAG and contextual benchmarks, such as HotpotQA and TriviaQA \citep{hotpotqa_yang2018,triviaqa_joshi2017,popqa_mallen2022not,truthfulqa_lin2021,wikimultihopqa_constructing_ho2020,musique_trivedi2022,natural_question_nq_kwiatkowski2019natural}, that standardizes the evaluation setup leading to consistent and reproducible evaluation results. In experiments, we show that our \ourmodel-9B model is both a well-rounded and high performing model, achieving state-of-the-art performance in three of the seven benchmarks in \oureval; see~\Cref{fig:radar-plot} for a preview of our results. \ourmodel-9B outperforms or is competitive with GPT-4o~\cite{gpt4} on all tasks in \oureval. It also outperforms powerful contextual models, such as Command-R+ \citep{commandr}, on a variety of tasks despite having 10 times fewer parameters. Compared to comparable baselines, our model is also shown to be resilient to factual alterations and unanswerability tests in the context. Lastly, despite being trained with a focus on RAG and contextual applications, our model is still competitive as a regular instruction-tuned LLM, with strong and comparable performances in standard benchmarks like MMLU or GSM8K \citep{mmlu_hendryckstest2021,gsm8k_cobbe2021gsm8k,arc_challenge_clark2018think}, as well as function-calling ability \citep{berkeley-function-calling-leaderboard}.

\section{\ourmodel}\label{sec:method}

In this section, we provide more insights into \ourmodel. First, we introduce a novel chat template comprising two new chat roles with specific functions (\Cref{sec:chat_template}). Then, we briefly discuss the training process of \ourmodel (\Cref{sec:sft_pref_learning}).

\subsection{\ourmodel Chat Template}\label{sec:chat_template}

\begin{figure}
\centering
\includegraphics[width=0.65\textwidth]{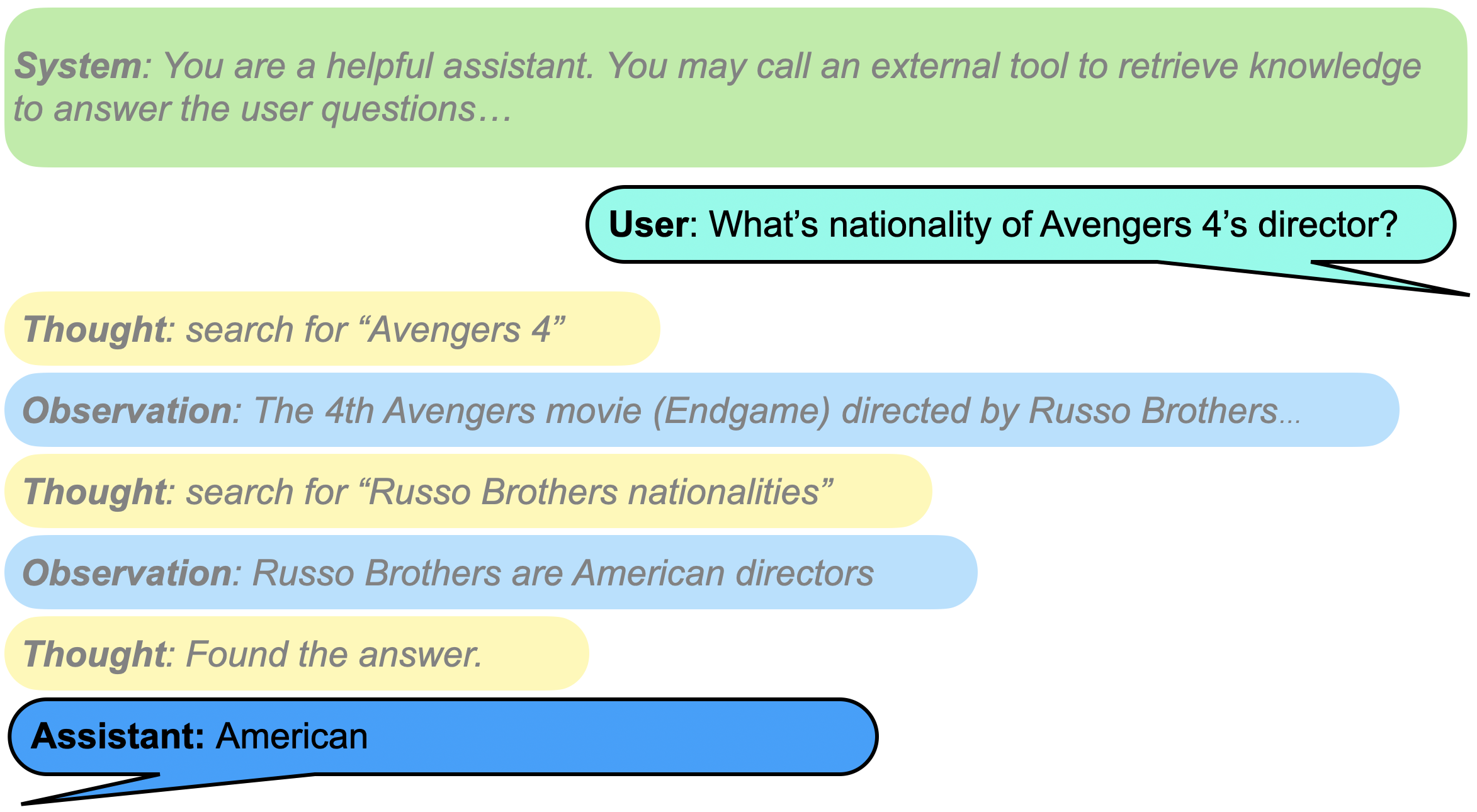}
\caption{Example of the chat format used by \ourmodel, with additional \thought and \observation turns (roles). The former indicates the model's ``inner'' thought or reasoning, actions and tool use syntax that are not typically meant to be shown to users. The latter indicates all external information retrieved and returned by performing a search or function call. The \assistant turn, therefore, is relieved to only be responsible to generate user-friendly responses. During training, \thought and \assistant turns are trained while the others are masked out.}
\label{fig:chat_template}
\end{figure}

Most instruction-tuned language models often feature a chat template that allows for three conversational roles: \Ni \system role, typically specified once at the beginning, is used to define the general characteristics of the AI assistant with  general instructions on how to respond to user inputs, \Nii \user role specifies where user messages reside, and \Niii \assistant turn is where the model responds to the user's query in accordance to the guidelines given by the \system turn. 

However, as more complex applications with (potentially multi-step) retrieval or function calling are being employed, such roles may have to handle increasingly complex and confusing data formats. For example, in retrieval tasks, external context information may be injected into the \system or \user turn, or may even form a part of the \assistant turn if the context is retrieved following a model's function call \citep{commandr,react_yao2022react}. This may cause confusion and distraction from the actual instruction or question queried by the user in the \user turn. In other words, there have been no generally agreed-upon position to store the contextual information in such tasks.
In another example of agentic function calling tasks, the \assistant turn has to produce responses that use specific tool syntax and expects to receive the function call's results following that \citep{react_yao2022react}. This makes the fine-tuning process tricky as the function's results, which are part of \assistant turn and typically contain answer clues, need to be masked out from the loss to prevent memorization \citep{llama2touvron2023llama}. Furthermore, for certain applications, practitioners may prefer to hide the model's reasoning, or intermediate actions invoked by the model from the user and only show user-friendly responses. Having all data processing steps enclosed within the \assistant turn may hinder such use cases.
Reliability is also an issue as the model may fail to produce specific key words consistently in inference strategies like ReAct \citep{react_yao2022react}.\footnote{ReAct uses arbitrarily literal phrases like ``Thought:'', ``Result:'' and ``Final Answer:'' within the \assistant turn to parse reasoning, tool outputs and answers respectively, which the LLM may not always comply with.}
Furthermore, the ambiguities in roles, functions and privileges may lead to the model failing to comply with the \system prompt and surrendering to jailbreaks injected via the \user turn or tool outputs \citep{instruction_hierarchy_wallace2024}.

To overcome such complexities, we introduce two more optional roles (turns) in the conversational template for \ourmodel: \thought and \observation. As shown in \Cref{fig:chat_template}, the \observation role is designated to house any contextual information acquired from external sources, which can be either retrieved documents for retrieval use case or function call's results in agentic tool use scenario. Meanwhile, the \thought role is designed for the model to speak out any internal reasoning, or tool use syntax to invoke certain function calls. The benefits of this change are manifold. First, it allows easy masking during training. Specifically, \system, \user and \observation turns may not be trainable as they are input information for the model to generate responses. Like \assistant, on the other hand, \thought turns are to be included in the fine-tuning loss to train the model to produce such ``thoughts''. Second, the separation and clarification of roles facilitate instruction hierarchy enforcement \citep{instruction_hierarchy_wallace2024}, making LLMs safer by ensuring them to respect the \system prompt and refuse to follow malicious instructions injected in \user and \observation turns. Third, the additional roles streamline the process of building reliable and secure RAG and agentic applications, by allowing developers to display or hide internal data processing steps, and avoid having to parse customized key words from the \assistant output.

\subsection{SFR-RAG Fine-tuning Process}\label{sec:sft_pref_learning}

One of the most important goals of \ourmodel is to make full use and complete comprehension of any provided contextual information in the real-world RAG scenarios. This trait includes many capabilities, among which are \Ni extracting relevant information from arbitrary long contexts, \Nii recognizing the lack of relevant information and abstaining from hallucinated generation, \Niii recognizing potential conflicting information in contextual passages, and \Niv being resilient to distracting, counter-intuitive information or contents that are out-of-distribution from the pre-training process. We fine-tuned SFR-RAG via standard supervised fine-tuning and preference learning \citep{llama2touvron2023llama,llama3_dubey2024llama,dpo_rafailov2023direct}, using extensive instruction-following data that mimic real-world retrieval question answering applications.

\section{Evaluation}\label{sec:eval}

\subsection{Contextual Evaluation Suite - \oureval}\label{sec:rag_evaluation}

There are already several evaluation protocols available to measure performance  of LLMs and RAG systems on contextual understanding across different domains and complexities \citep{triviaqa_joshi2017,truthfulqa_lin2021,popqa_mallen2022not,hotpotqa_yang2018,wikimultihopqa_constructing_ho2020,musique_trivedi2022,arc_challenge_clark2018think,natural_question_nq_kwiatkowski2019natural}. However, prior studies \citep{commandr,rag2,selfrag_asai2023self,react_yao2022react} have reported results on non-overlapping measures, datasets and inconsistent setups, especially on which contextual content to present to the LLMs and model hyper-parameters. This causes challenges in directly comparing results from different studies. 

To offer a better common ground, we propose \textbf{\oureval}, which is primarily an aggregation of 7 popular contextual question answering tasks, namely HotpotQA, TriviaQA, TruthfulQA, PopQA, 2WikiHopQA, Musique and Natural Questions (NQ) \citep{triviaqa_joshi2017,truthfulqa_lin2021,popqa_mallen2022not,hotpotqa_yang2018,wikimultihopqa_constructing_ho2020,musique_trivedi2022,natural_question_nq_kwiatkowski2019natural}. There are a few key contributions that make \oureval stand out as a comprehensive benchmarking framework for contextual LLMs:

\begin{itemize}[leftmargin=*]
    \item The measures in \oureval are evaluated under the same instruction with contextual contents being consistently specified. The contextual contents include the original context documents of each benchmark if provided, or otherwise they are retrieved from a much larger Wikipedia database with an embedding model of choice.
    \item As it is not trivial to evaluate assistant-style LLMs with a wide range of verbosity in the generated output, \oureval offers multiple scoring methods to account for variations in answers compared to the ground truths. These scoring methods are \Ni Exact Match (EM) of the generated answer with the ground truth, \Nii Easy match (EasyM) to check if the ground truth is in the generated answer, and \Niii the F1 score, with room for further addition. The Appendix provides the complete results for all of these metrics.
    \item \oureval offers multiple setups common in the RAG scenarios, among which is whether to retrieve top-k chunks using consistent embedding models or feed the entire available contextual documents directly to the LLM (no retrieval needed). 
\end{itemize}

The variety of measures and tasks in \oureval enables both holistic and specific evaluation of contextual LLMs. By weighing each task and measure equally, \oureval allows for direct comparison of model performance in general. On the other hand, depending on practitioner use-case and domain specifications, certain measures or datasets may be prioritized, allowing for quick identification of the best task-specific models.

\paragraph{Dataset Specific Settings.}

For 2WikiHopQA, HotpotQA and Musique, the context documents are already provided for each question, so we use them directly as contextual sources. For TriviaQA, TruthfulQA, and NQ, the questions come with their respective Wikipedia article or source URL. We scraped the web content from these sources and used Cohere embedding \citep{cohere_embedding} to retrieve top-10 chunks from the contextual sources where each chunk is 512 tokens long.  Meanwhile, PopQA itself does not come with context documents, so we make use of the off-the-shelf context documents produced by the Self-RAG retriever \citep{selfrag_asai2023self}. For each task, we use the test set if they are complete with gold labels, otherwise we use the entire validation set to measure models' performance. This is different from Command-R's report \citep{commandr}, where HotpotQA evaluation was conducted on a 100-sample subset of validation set, with no details about the context documents disclosed.

Note that \oureval contains popular existing benchmarks, such as TriviaQA and TruthfulQA, where evaluation utilizes certain contexts to which models are expected to be faithful. That is, models are expected to utilize only the information found in such contexts, in contrast to traditional closed-book QA settings, where the parametric knowledge of LLMs are evaluated sans provided contexts. In other words, the presence of these contexts may cause the scores to differ significantly.

\subsection{Experimental Results on \oureval}

\begin{table}[t]
  \caption{Performances of \ourmodel-9B and various open- and closed-source baselines across 7 contextual question answering tasks in \oureval. Bold numbers mean best of all, while underlined numbers mean best among open-source models. PopQA is measured in easy-match accuracy (EasyEM), while the rest are measured in exact-match accuracy (EM). The Appendix presents the full results in metrics.}
  \label{table:eval:contextual_bench}
  \centering
  \resizebox{\textwidth}{!}{%
  \setlength{\tabcolsep}{5pt}
  \begin{tabular}{lccccccc|c}
    \toprule
    {\bf Model} & {\bf TriviaQA} & {\bf TruthfulQA} & {\bf 2WikiHopQA} & {\bf MuSiQue} & {\bf NQ} & {\bf PopQA} & {\bf HotpotQA}  & {\bf Average}  \\
    \midrule
    GPT-4o                 & \textbf{81.73} & 76.47 & 62.40 & \textbf{44.40} & \textbf{51.85} & 42.24             & 57.20  & 59.47 \\
    GPT 4 Turbo            & 78.34          & 76.13 & 59.90 & 37.10          & 48.23          & 53.82                      & 54.20  & 58.25 \\
    GPT-4o-mini            & 72.55          & 61.02 & 46.60 & 32.10          & 30.75          & 54.75                      & 49.80  & 49.65 \\
    GPT 3.5 Turbo          & 77.43          & 47.00 & 41.20 & 22.60          & 40.40          & 53.39                      & 48.10  & 47.16 \\
    \midrule
    Command R (35B)        & 71.75          & 51.65 & 41.50 & 14.60          & 35.76          & 52.60                      & 48.20  & 45.15 \\
    Command R+ (104B)      & 76.34          & 56.30 & 47.50 & 29.00          & 47.44          & 59.18                      & 52.50  & 52.61    \\
    gemma-2-9b-it          & 77.66          & 59.43 & 49.70 & 30.50          & 37.37          & 53.75                      & 52.60  & 51.57 \\
    Llama3 8B Instruct     & 72.28          & 51.52 & 27.90 & 9.90           & 31.01          & 52.82                      & 45.90  & 41.62 \\
    Llama3.1 8B Instruct   & 71.33          & 59.43 & 17.70 & 9.40           & 44.41          & \textbf{\underline{59.68}} & 46.70  & 44.09 \\
    \midrule
    \ourmodel-9B           & \underline{79.24}   & \underline{\textbf{77.45}} & \textbf{\underline{79.50}} & \underline{42.90}          & \underline{48.01}          & 53.97                      & \underline{\textbf{65.70}}  & \underline{\textbf{63.44}} \\
    \bottomrule
  \end{tabular}
  }
\end{table}

\Cref{table:eval:contextual_bench} compares the performance of our 9B \ourmodel model on \oureval against state-of-the-art large models as well as comparable ones across the 7 question answering tasks. PopQA scores are measured in easy matching, while the remaining are measured in exact matching. As shown, GPT-4o \citep{gpt4} unsurprisingly aces most of the benchmarks. However, given its small size, our \ourmodel-9B model significantly outperforms strong open-source baselines such as Command-R and Command-R+ that have up to 10 times larger parameter counts. Remarkably, it achieves the state of the art in TruthfulQA, 2WikihopQA and HotpotQA in contextual settings. Overall, it also achieves the state of the art average performance, demonstrating our model's strong ability across many contextual tasks. In particular, our model excels at 2WikiHopQA, with nearly a 25\% increase in performance compared to GPT-4o.
Meanwhile, our 9B model consistently outperforms Llama-3.1 8B Instruct and gemma-2-9b-it across most benchmarks.

\begin{figure}[h]
\centering
\begin{tikzpicture}  
\begin{axis}  
[  
    ybar, 
    enlargelimits=0.15,
    legend style={at={(0.5,-0.25)}, 
      anchor=north,legend columns=-1},     
    width=\textwidth,
    height=0.4\textwidth,
    ylabel={EasyM Accuracy}, 
    symbolic x coords={Command-R+, GPT-4o, gemma-2-9b-it, SFR-RAG-9B},  
    xtick=data,  
    nodes near coords,  
    bar width=15pt,
    every node near coord/.append style={font=\small},
    nodes near coords align={vertical},  
    ]  
\addplot coordinates {(Command-R+, 73.6) (GPT-4o, 47.2) (gemma-2-9b-it, 55.7) (SFR-RAG-9B, 85.8) };  
\addplot coordinates {(Command-R+, 40.0) (GPT-4o, 59.9) (gemma-2-9b-it, 48.7) (SFR-RAG-9B, 70.6) };  
\addplot coordinates {(Command-R+, 56.6) (GPT-4o, 93.2) (gemma-2-9b-it, 50.7) (SFR-RAG-9B, 99.2) }; 

\legend{Counterfactual, Unknown, Conflict}  
\end{axis}  
\end{tikzpicture}
\caption{FaithEval \citep{faitheval}: average easy match accuracy scores of different models when contextual facts are fabricated (Counterfactual), removed (Unknown) or when the facts are contradicting (Conflict). Small variations between those settings and overall high absolute scores indicate that \ourmodel-9B is resilient to changes in contextual information.}
\label{fig:faitheval}
\end{figure}
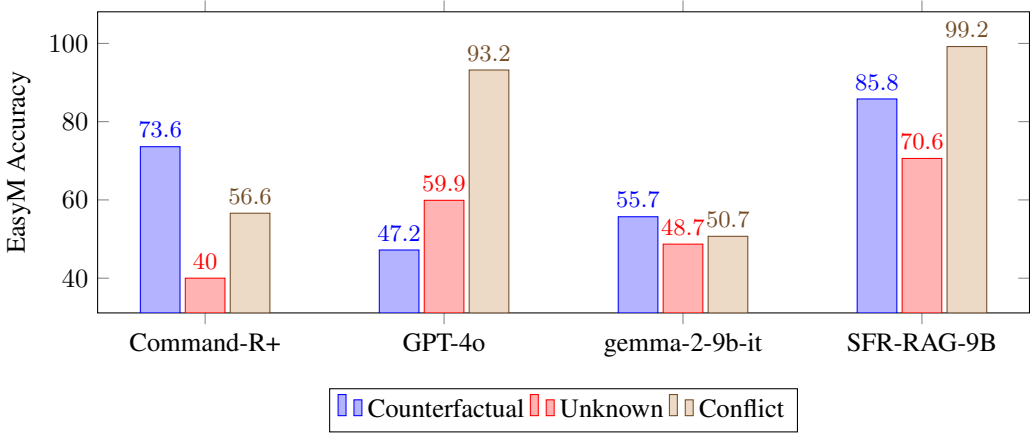

\subsection{Resilience to Unanswerable, Conflicting and Counterfactual Contexts}\label{sec:faithe}

Because most QA benchmarks are realistically based on real-world facts, understanding LLMs performance in contextual QA tasks may be ambiguous because high scores may be attributed to either \Ni the ability to seek accurate facts from the contextual documents and content, or \Nii the intrinsic parametric knowledge of the model acquired during pre-training, of which large and state-of-the-art models like GPT-4o often have significant advantages.

\citet{faitheval} recently proposed \textbf{FaithEval}, an evaluation suite that measures how LLMs remain faithful to the context if the facts of the contexts are changed. The benchmark evaluate LLMs on three  scenarios: \Ni ``Unknown'' where the relevant facts are removed and the original question becomes unanswerable; \Nii ``Conflict'' where multiple context documents are provided that contains conflicting or contradicting information and the model is expected to recognize that; and \Niii ``Counterfactual'' where certain commonsense facts are altered by introducing a falsely fabricated context document. For instance, ``The Moon is Made of Marshmallows.'' is considered a \emph{counterfactual} context and the  LLM under evaluation is expected to remain faithful to that ``fact'', if prompted to do so. Following \citet{faitheval}, the ``Unknown'' and ``Conflict'' tasks are averaged over 10 benchmarks \citep{bioasq_tsatsaronis2012,drop_dua2019drop,hotpotqa_yang2018,natural_question_nq_kwiatkowski2019natural,newsqa_trischler2016,raceqa-lai-etal-2017-race,squad_rajpurkar2016,searchqa_dunn2017,textbookqa_kembhavi2017you,triviaqa_joshi2017}, while the ``Counterfactual'' task is evaluated using the ARC-C dataset \citep{arc_challenge_clark2018think}.

\begin{table}[h]
  \caption{Standard LM-eval-harness benchmarks \citep{lm-eval-harness}: \ourmodel-9B maintains relative competitiveness in standard world knowledge and reasoning abilities against comparable baselines.}
  \label{table:eval:standard_benchmark}
  \centering
  \resizebox{\textwidth}{!}{%
  \begin{tabular}{lcccccc}
    \toprule
    {\bf Model} &  {\bf MMLU } & {\bf GSM8K } & {\bf Winogrande } & {\bf TruthfulQA } & {\bf Hellaswag } & {\bf ARC-C } \\
    \midrule
    Command-R (35B) & 68.20 & 56.63 & 81.53 & 52.32 & 87.00 & 65.53 \\
    Llama-3-8b-instruct  & 67.07 & 68.69 & 74.51 & 51.65 & 78.55 & 60.75\\
    Llama-3.1-8B-Instruct  & 68.22 & 71.04 & 78.06 & 54.58 & 80.47 & 60.92\\
    gemma-2-9b-it & 70.80 & 76.88 & 77.50 & 60.11 & 81.78 & 71.20\\
    \midrule
    \ourmodel-9B     &  70.15 & 82.56 & 78.46 & 56.49 & 81.58 & 69.12 \\
    \bottomrule
  \end{tabular}
  }
\end{table}

\begin{table}[h]
  \caption{Scores on Berkeley function calling benchmark \citep{berkeley-function-calling-leaderboard}: \ourmodel-9B exhibits competitive function calling abilities.}
  \label{table:eval:function_calling}
  \centering
  \begin{tabular}{lccc}
    \toprule
    {\bf Model} &  {\bf Executable } & {\bf AST } & {\bf  Relevance }\\
    \midrule
    GPT 4 Turbo	& 86.04 & 90.73 & 	62.50 \\
    GPT 3.5 Turbo & 81.38 & 75.23 & 87.80 \\
    Command R + (104B) & 77.33 & 	84.50 & 	63.75 \\
    xLam 7B	& 87.12 & 	89.46 & 	85.00 \\
    Mistral Medium & 73.47 & 	84.48 & 	88.33 \\
    gemma-2-9b-it & 70.50 & 	69.69 & 	90.00 \\
    Meta-Llama-3-8B-Instruct & 65.13 & 	61.63 & 	26.60 \\
    Meta-Llama-3.1-8B-Instruct & 75.17 & 	77.56 & 	40.00 \\
    \midrule
    \ourmodel-9B &  70.88	& 71.69	& 72.50 \\
    \bottomrule
  \end{tabular}
\end{table}

\Cref{fig:faitheval} shows the average non-strict matching accuracy scores of different LLMs over 3 tasks under FaithEval suite.
As shown, other baselines such as GPT-4o exhibit high variations when the facts change in Counterfactual and Unknown settings. Particularly, GPT-4o scores low in Counterfactual setting perhaps because large models may have higher knowledge inertia and stronger resistance to factual changes.
Meanwhile, our \ourmodel-9B scores consistently highly, even when the context information is altered. This demonstrates that our model is usefully resilient and faithful to unseen contextual information. It also means that the model is more adaptable to the ever-changing world. Plus, our model is more capable of identifying contradiction in the contexts, as well as resisting against its own parametric knowledge when contextual information presented is counter-intuitive. In other words, the model remains more faithful to the context even if the context contradicts its pre-trained knowledge.

\subsection{Standard Benchmarks}\label{sec:standard_benchmarks}

We also evaluate our \ourmodel model in the traditional few-shot prompting benchmarks \citep{mmlu_hendryckstest2021,gsm8k_cobbe2021gsm8k} to measure its parametric knowledge as well as general instruction following and reasoning abilities. Using the similar setups in the Open LLM leaderboard \citep{open-llm-leaderboard}, we employ the standard evaluation harness \citep{lm-eval-harness} to evaluate our model in MMLU (5 shots), GSM8K (5 shots with strict matching), Winogrande (5 shots), TruthfulQA (0 shot MC2), Hellaswag (10 shots with normalized accuracy) and ARC-C (25 shots with normalized accuracy) \citep{mmlu_hendryckstest2021,gsm8k_cobbe2021gsm8k,winogrande_sakaguchi2021,truthfulqa_lin2021,hellaswag_zellers2019,arc_challenge_clark2018think}.

As shown in \Cref{table:eval:standard_benchmark}, our \ourmodel model performs competitively in terms of world knowledge, common sense and reasoning abilities, despite the fact that it is optimized for contextual and retrieval use cases.
Particularly, our 9B model outperforms Command-R \citep{commandr} with 35B parameters in MMLU, GSM8K, TruthfulQA as well as ARC-C. Meanwhile it remains competitive to Llama-3.1-Instruct \citep{llama3_dubey2024llama} and gemma-2-9b-it \citep{gemma_team2024}.

Our model is also trained with function calling with a focus to support dynamic and multi-hop interactions with external tools to retrieve high-quality contextual information. As such, we compare our model with certain popular baselines in the Berkeley function calling task \citep{berkeley-function-calling-leaderboard}. As reported in \Cref{table:eval:function_calling}, our \ourmodel performs competitively against comparable baselines such as Llama-3-8B-Instruct \cite{llama3_dubey2024llama}

\section{Conclusion}

We present \ourmodel, a LLM fine-tuned with an emphasis on faithful contextual comprehension and understanding for retrieval augmented generation applications. The \ourmodel model is trained to minimize hallucination, effectively handle unanswerable, counterfactual or low-quality and irrelevant contexts. It is also capable of performing complex multi-hop reasoning and producing citations reliably and accurately. We also introduce ContextualBench, which is a compilation of various popular RAG benchmarks evaluated under consistent and appropriate settings. The experiments show that our 9B \ourmodel model outperforms notable baselines, including Command-R+ and GPT-4o, and achieves the state of the art in 3 out of 7 benchmarks in ContextualBench. Our evaluation with FaithEval also shows that our model is resilient to changes in the context information and able to identify unanswerable questions. Lastly, the \ourmodel model maintains competitive performance in general instruction-tuned tasks and function-calling capabilities, compared to baselines of similar sizes.

\bibliography{neurips_2023.bib}
\bibliographystyle{plainnat}

\newpage
\appendix

\section{Appendix}


\begin{table}[h]
  \caption{\oureval scores for HotpotQA, 2WikihopQA and Musique, as measured in easy matching, strict matching accuracy and the F1 score.}
  \label{table:eval:hotpot_wiki_musique}
  \centering
  \resizebox{\textwidth}{!}{%
  \setlength{\tabcolsep}{5pt}
  \begin{tabular}{lccccccccc}
    \toprule
    \multirow{2}{*}{\bf Model} & \multicolumn{3}{c}{\bf HotpotQA} & \multicolumn{3}{c}{\bf 2WikiHopQA} & \multicolumn{3}{c}{\bf MuSiQue} \\
    \cmidrule{2-10}
                                & EasyM & EM & F1 & EasyM & EM & F1 & EasyM & EM & F1 \\
    \midrule
    GPT 4 Turbo & 70.90 & 54.20 & 69.00 & 73.00 & 59.90 & 69.60 & 53.10 & 37.10 & 52.70 \\
    GPT 4o & 72.70 & 57.2 & 71.40 & 74.70 & 62.40 & 70.90 & 60.20 & 44.40 & 58.60 \\
    GPT-4o-mini & 68.70 & 49.80 & 65.50 & 57.70 & 46.60 & 56.50 & 48.30 & 32.10 & 48.90 \\
    GPT 3.5 Turbo & 63.10 & 48.10 & 61.80 & 50.50 & 41.20 & 49.70 & 34.90 & 22.60 & 35.90 \\
    \midrule
    Command R (35B) & 63.10 & 48.20 & 63.00 & 55.70 & 41.50 & 52.30 & 36.70 & 14.60 & 34.10 \\
    Command R + (104B) & 67.30 & 52.50 & 66.60 & 60.80 & 47.50 & 57.30 & 46.30 & 29.00 & 43.60 \\
    LLama3 8B Instruct (8B) & 58.80 & 45.90 & 60.00 & 46.50 & 27.90 & 42.20 & 29.30 & 9.90 & 25.30 \\
    LLama3.1 8B Instruct (8B) & 64.80 & 46.70 & 62.00 & 51.70 & 17.70 & 38.30 & 38.00 & 9.40 & 29.00 \\
    gemma-2-9b-it & 67.20 & 52.60 & 66.10 & 59.60 & 49.70 & 58.10 & 47.00 & 30.50 & 43.50 \\
    \midrule
    \ourmodel-9B & 83.60 & 65.70 & 79.90 & 86.20 & 79.50 & 84.60 & 57.40 & 42.90 & 53.50\\
    \bottomrule
  \end{tabular}
  }
\end{table}

\begin{table}[h]
  \caption{\oureval scores for TriviaQA, TruthfulQA, PopQA and NQ, as measured in easy matching, strict matching accuracy and the F1 score.}
  \label{table:eval:trivia_truthful_popqa_nq}
  \centering
  \resizebox{\textwidth}{!}{%
  \setlength{\tabcolsep}{5pt}
  \begin{tabular}{lcccccc}
    \toprule
    \multirow{2}{*}{\bf Model} & \multicolumn{2}{c}{\bf TriviaQA} & {\bf TruthfulQA} & {\bf PopQA} & \multicolumn{2}{c}{\bf NQ}\\
    \cmidrule{2-7}
                                & EM & F1 & EM & EasyM & EM & F1 \\
    \midrule
GPT 4 Turbo   & 78.34 & 86.30 & 76.13 & 53.82 & 48.23 & 67.20 \\
GPT 4o   & 81.73 & 88.20 & 76.47 & 42.24 & 51.85 & 69.75 \\
GPT-4o-mini   & 72.55 & 81.77 & 61.02 & 54.75 & 30.75 & 54.90 \\
GPT 3.5 Turbo   & 77.43 & 84.70 & 47.00 & 53.39 & 40.40 & 60.69 \\
\midrule
Command R (35B)   & 71.75 & 79.81 & 51.65 & 52.60 & 35.76 & 55.50 \\
Command R + (104B)   & 76.34 & 83.05 & 56.30 & 59.18 & 47.44 & 63.21 \\
LLama3 8B Instruct (8B)   & 72.28 & 79.07 & 51.52 & 52.82 & 31.01 & 48.22 \\
LLama3.1 8B Instruct (8B)   & 71.33 & 78.33 & 59.43 & 59.68 & 44.41 & 61.60 \\
gemma-2-9b-it   & 77.66 & 83.23 & 59.43 & 53.75 & 37.37 & 53.15 \\
    \midrule
    \ourmodel-9B & 79.24 & 84.35 & 77.45 & 53.97 & 48.01 & 59.66\\
    \bottomrule
  \end{tabular}
  }
\end{table}

\Cref{table:eval:hotpot_wiki_musique,table:eval:trivia_truthful_popqa_nq} present the complete results of \oureval across 7 benchmarks as measured in different metrics. As these metrics measure different aspects of the question-answering responses, the numbers provide more insights into the performance comparison between for models. For instance, a model that is too verbose may score low in EM and high in F1, and vice-versa.

\end{document}